\documentclass[10pt,twocolumn,letterpaper]{article}

\usepackage{iccv}
\usepackage{times}
\usepackage{epsfig}
\usepackage{graphicx}
\usepackage{amsmath}
\usepackage{amssymb}
\usepackage{subfigure}
\usepackage{color}
\usepackage{multirow}
\usepackage{cite}
\usepackage{epstopdf}
\usepackage{pifont}
\usepackage[dvipsnames]{xcolor}
\usepackage{authblk}
\usepackage{fancyhdr}
\def\figvspace{{\vspace{-5mm}}}
\newcommand{\Paragraph}[1]{\vspace{-0mm} \noindent \textbf{#1} \hspace{0mm}}
\newcommand{\Section}[1]{\vspace{-1mm} \section{#1} \vspace{-1mm}}
\newcommand{\SubSection}[1]{\vspace{-1mm} \subsection{#1} \vspace{-1mm}}

\newcommand*{\boxedcolor}{red}
\makeatletter
\renewcommand{\boxed}[1]{\textcolor{\boxedcolor}{%
  \fbox{\normalcolor\m@th$\displaystyle#1$}}}
\makeatother

\makeatletter
  \newcommand\figcaption{\def\@captype{figure}\caption}
  \newcommand\tabcaption{\def\@captype{table}\caption}
\makeatother

\usepackage[breaklinks=true,bookmarks=false]{hyperref}

\iccvfinalcopy 

\def\iccvPaperID{****} 

\ificcvfinal\fi

\begin{document}

\title{MemNet: A Persistent Memory Network for Image Restoration}

\author[1]{Ying Tai\thanks{This work was supported by the National Science Fund of China under Grant Nos. $91420201$, $61472187$, $61502235$, $61233011$, $61373063$ and $61602244$, the $973$ Program No. $2014$CB$349303$, Program for Changjiang Scholars, and partially sponsored by CCF-Tencent Open Research Fund. Jian Yang and Xiaoming Liu are corresponding authors.} }
\author[1]{Jian Yang}
\author[2]{Xiaoming Liu}
\author[1]{Chunyan Xu}

\affil[1]{Department of Computer Science and Engineering, Nanjing University of Science and Technology}
\affil[2]{Department of Computer Science and Engineering, Michigan State University
\authorcr \texttt{\small\{taiying, csjyang, cyx\}@njust.edu.cn, liuxm@cse.msu.edu}}

\maketitle
\thispagestyle{empty}

\begin{abstract}
   Recently, very deep convolutional neural networks (CNNs) have been attracting considerable attention in image restoration.
   However, as the depth grows, the long-term dependency problem is rarely realized for these very deep models, which results in the prior states/layers having little influence on the subsequent ones.
   Motivated by the fact that human thoughts have persistency, we propose a very deep persistent memory network (MemNet) that introduces a memory block, consisting of a recursive unit and a gate unit, to explicitly mine persistent memory through an adaptive learning process.
   The recursive unit learns multi-level representations of the current state under different receptive fields.
   The representations and the outputs from the previous memory blocks are concatenated and sent to the gate unit, which adaptively controls how much of the previous states should be reserved, and decides how much of the current state should be stored.
   We apply MemNet to three image restoration tasks, i.e., image denosing, super-resolution and JPEG deblocking.
   Comprehensive experiments demonstrate the necessity of the MemNet and its unanimous superiority on all three tasks over the state of the arts.
   Code is available at~\url{https://github.com/tyshiwo/MemNet}.
\end{abstract}

\begin{figure}[t!]
  \centering
  \includegraphics[trim={0 0 0 0mm},clip,width=0.75\linewidth]{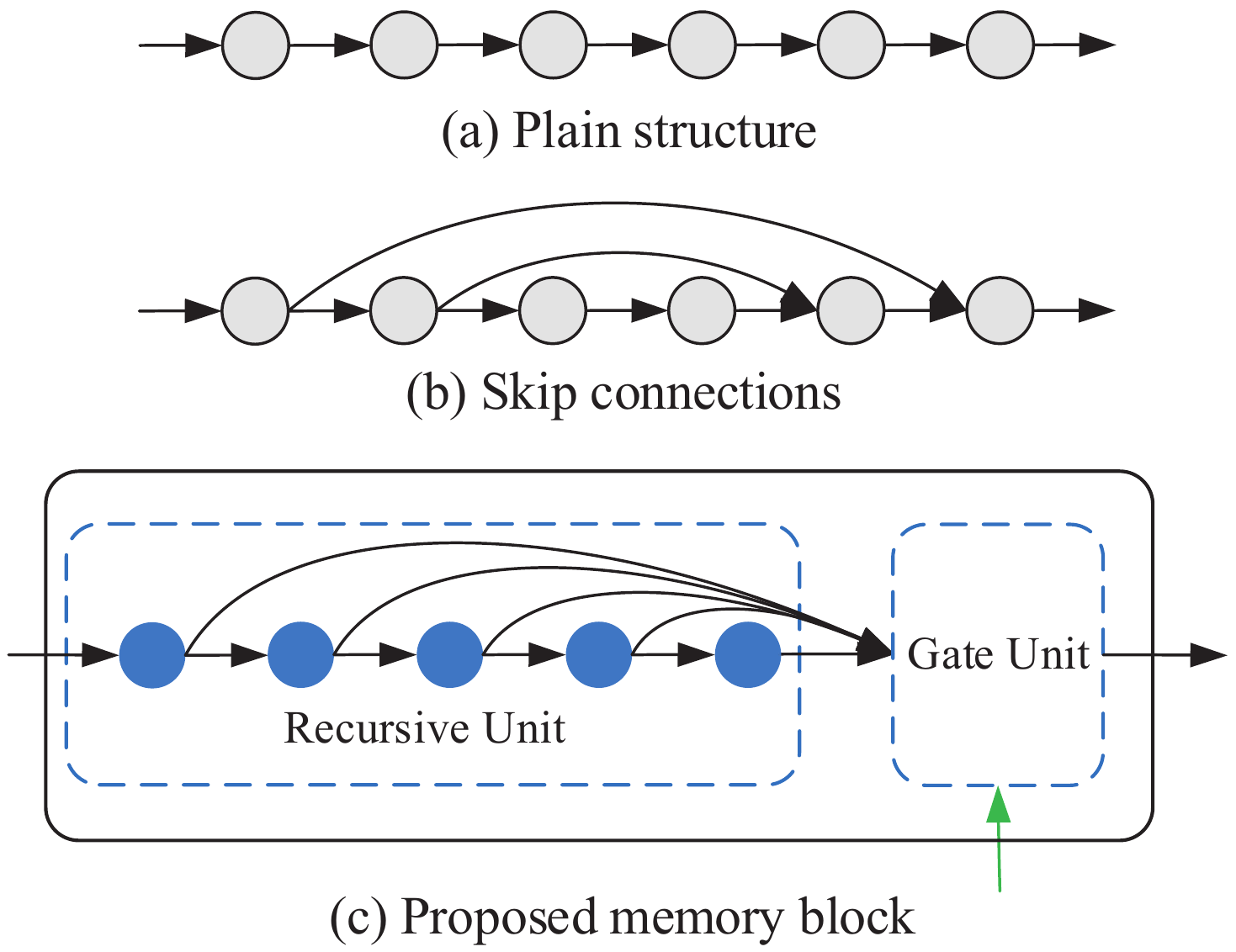}
  \caption{\small Prior network structures (a,b) and our memory block (c). The \textcolor{blue}{blue circles} denote a recursive unit with an unfolded structure which generates the short-term memory.
  The \textcolor{green}{green arrow} denotes the long-term memory from the previous memory blocks that is directly passed to the gate unit.}
  \label{fig:Differences} \figvspace
\end{figure}

\Section{Introduction}\label{section:1}
Image restoration~\cite{IR_survey} is a classical problem in low-level computer vision, which estimates an uncorrupted image from a noisy or blurry one.
 A corrupted low-quality image $\mathbf{x}$ can be represented as: $\mathbf{x}=D(\tilde{\mathbf{x}})+\mathbf{n}$, where $\tilde{\mathbf{x}}$ is a high-quality version of $\mathbf{x}$, $D$ is the degradation function and $\mathbf{n}$ is the additive noise.
With this mathematical model, extensive studies are conducted on many image restoration tasks, e.g., image denoising~\cite{BM3D,PCLR,PGPD,WNNM}, single-image super-resolution (SISR)~\cite{ScSR,SelfESR_CVPR15} and JPEG deblocking~\cite{JP02_ECCV12,DSC_CVPR15}.

As three classical image restoration tasks, image denoising aims to recover a clean image from a noisy observation, which commonly assumes additive white Gaussian noise with a standard deviation $\sigma$;
single-image super-resolution recovers a high-resolution (HR) image from a low-resolution (LR) image;
and JPEG deblocking removes the blocking artifact caused by JPEG compression~\cite{ARCNN_ICCV15}.

Recently, due to the powerful learning ability, very deep convolutional neural network (CNN) is widely used to tackle the image restoration tasks.
Kim~et~al.~construct a $20$-layer CNN structure named VDSR for SISR~\cite{VDSR_CVPR16}, and adopts residual learning to ease training difficulty.
To control the parameter number of very deep models, the authors further introduce a recursive layer and propose a Deeply-Recursive Convolutional Network (DRCN)~\cite{DRCN_CVPR16}.
To mitegate training difficulty, Mao~et~al.~\cite{RED_NIPS16} introduce symmetric skip connections into a $30$-layer convolutional auto-encoder network named RED for image denoising and SISR.
Moreover, Zhang~et~al.~\cite{DnCNN} propose a denoising convolutional neural network (DnCNN) to tackle image denoising, SISR and JPEG deblocking simultaneously.

The conventional plain CNNs, e.g.,~VDSR~\cite{VDSR_CVPR16}, DRCN~\cite{DRCN_CVPR16} and DnCNN~\cite{DnCNN} (Fig.~\ref{fig:Differences}(a)), adopt the single-path feed-forward architecture, where one state is mainly influenced by its direct former state, namely \textit{short-term memory}.
Some variants of CNNs, RED~\cite{RED_NIPS16} and ResNet~\cite{ResNet_CVPR16} (Fig.~\ref{fig:Differences}(b)), have skip connections to pass information across several layers.
In these networks, apart from the short-term memory, one state is also influenced by a \textit{specific} prior state, namely \textit{restricted long-term memory}.
In essence, recent evidence suggests that mammalian brain may protect previously-acquired knowledge in neocortical circuits~\cite{pesistent_nature_2015}.
However, none of above CNN models has such mechanism to achieve persistent memory.
As the depth grows, they face the issue of lacking long-term memory. 

To address this issue,
we propose a very deep persistent memory network (MemNet), which introduces a memory block 
to explicitly mine persistent memory through an adaptive learning process.
In MemNet, a Feature Extraction Net (FENet) first extracts features from the low-quality image.
Then, several memory blocks are stacked with a \textit{densely connected structure} to solve the image restoration task.
Finally, a Reconstruction Net (ReconNet) is adopted to learn the residual, rather than the direct mapping, to ease the training difficulty.

As the key component of MemNet, a memory block contains a \textit{recursive unit} and a \textit{gate unit}.
Inspired by neuroscience~\cite{Neuroscience,RCNN_CVPR15} that recursive connections ubiquitously exist in the neocortex, the recursive unit learns \textit{multi-level} representations of the current state under different receptive fields (\textcolor{blue}{blue circles} in Fig.~\ref{fig:Differences}(c)), which can be seen as the \textit{short-term memory}.
The \textit{short-term memory} generated from the recursive unit, and the \textit{long-term memory} generated from the previous memory blocks~\footnote{For the first memory block, its long-term memory comes from the output of FENet.} (\textcolor{green}{green arrow} in Fig.~\ref{fig:Differences}(c)) are \textit{concatenated} and sent to the gate unit, which is a non-linear function to maintain persistent memory.
Further, we present an extended multi-supervised MemNet, which fuses all intermediate predictions of memory blocks to boost the performance.

In summary, the main contributions of this work include:

$\diamond$ A memory block to accomplish the \textit{gating mechanism} to help bridge the long-term dependencies.
In each memory block, the gate unit adaptively learns different weights for different memories, which controls how much of the long-term memory should be reserved, and decides how much of the short-term memory should be stored.

$\diamond$ A very deep end-to-end persistent memory network ($\textit{80}$ convolutional layers)
for image restoration.
The densely connected structure helps compensate mid/high-frequency signals, and ensures maximum information flow between memory blocks as well.
To the best of our knowledge, it is by far the \textit{deepest} network for image restoration.

$\diamond$ The \textit{same} MemNet structure achieves the state-of-the-art performance in image denoising, super-resolution and JPEG deblocking.
Due to the strong learning ability, our MemNet can be trained to handle different levels of corruption even using a \textit{single} model.


\begin{figure*}[tbp]
  \centering
  \includegraphics[trim={0 0 0 0mm},clip,width=0.87\linewidth]{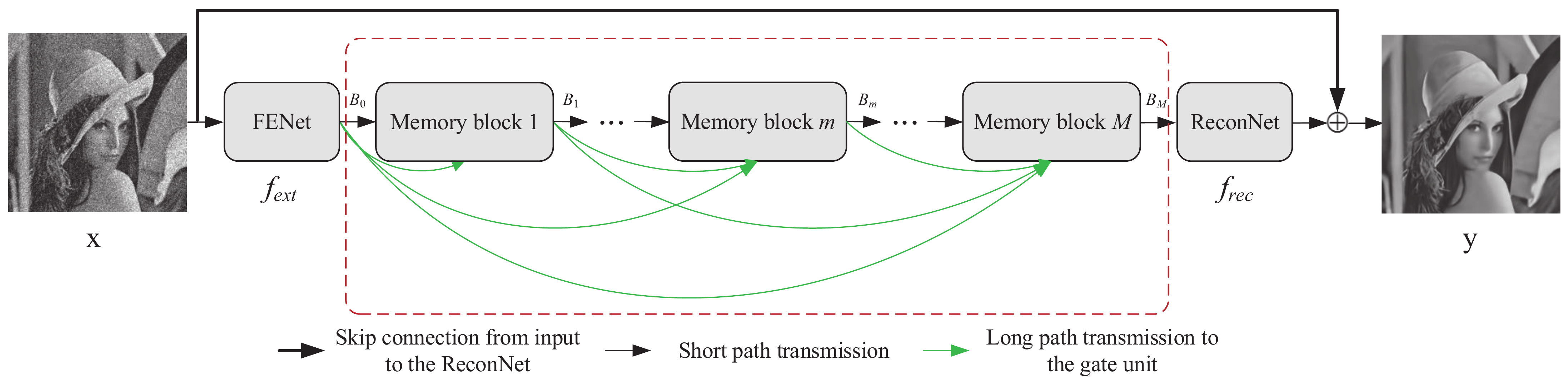}
  \caption{\small Basic MemNet architecture.
  The \textcolor{Red}{red dashed box} represents multiple stacked memory blocks.}
  \label{fig:NetworkStructure} \figvspace
\end{figure*}

\Section{Related Work}\label{section:2}



The success of AlexNet~\cite{AlexNet_NIPS12} in ImageNet~\cite{ImageNet} starts the era of deep learning for vision, and the popular networks, GoogleNet~\cite{GoogleNet_CVPR15}, Highway network~\cite{Highway}, ResNet~\cite{ResNet_CVPR16}, reveal that the network depth is of crucial importance.

As the early attempt, Jain et al.~\cite{CNNID_NIPS08} proposed a simple CNN to recover a clean natural image from a noisy observation and achieved comparable performance with the wavelet methods.
As the pioneer CNN model for SISR, super-resolution convolutional neural network (SRCNN)~\cite{SRCNN_PAMI16} predicts the nonlinear LR-HR mapping via a fully deep convolutional network, which significantly outperforms classical shallow methods.
The authors further proposed an extended CNN model, named Artifacts Reduction Convolutional Neural Networks (ARCNN)~\cite{ARCNN_ICCV15}, to effectively handle JPEG compression artifacts.

To incorporate task-specific priors, Wang et al.~adopted a cascaded sparse coding network to fully exploit the natural sparsity of images~\cite{CSCN_ICCV15}.
In~\cite{D3_CVPR16}, a deep dual-domain approach is proposed to combine both the prior knowledge in the JPEG compression scheme and the practice of dual-domain sparse coding.
Guo et al.~\cite{DDCN_ECCV16} also proposed a dual-domain convolutional network that jointly learns a very deep network in both DCT and pixel domains.

Recently, 
very deep CNNs become popular for image restoration.
Kim et al.\cite{VDSR_CVPR16} stacked $20$ convolutional layers to exploit large contextual information.
Residual learning and adjustable gradient clipping are used to speed up the training. 
Zhang et al.~\cite{DnCNN} introduced batch normalization into a DnCNN model to jointly handle several image restoration tasks.
To reduce the model complexity, the DRCN model introduced recursive-supervision and skip-connection to mitigate the training difficulty~\cite{DRCN_CVPR16}. 
Using symmetric skip connections, Mao et al.~\cite{RED_NIPS16} proposed a very deep convolutional auto-encoder network for image denoising and SISR.
Very Recently, Lai~et~al.~\cite{LapSRN} proposed LapSRN to address the problems of speed and accuracy for SISR, which operates on LR images directly and progressively reconstruct the sub-band residuals of HR images.
Tai~et~al.~\cite{DRRN} proposed deep recursive residual network (DRRN) to address the problems of model parameters and accuracy, which recursively learns the residual unit in a multi-path model.

\Section{MemNet for Image Restoration}\label{section:3}

\SubSection{Basic Network Architecture}\label{section:3.1}
Our MemNet consists of three parts: a feature extraction net FENet, multiple stacked memory blocks and finally a reconstruction net ReconNet (Fig.~\ref{fig:NetworkStructure}).
Let's denote $\mathbf{x}$ and $\mathbf{y}$ as the input and output of MemNet.
Specifically, a convolutional layer is used in FENet to extract the features from the noisy or blurry input image,
\begin{equation}\
  B_0 =  f_{ext}(\mathbf{x}),
  \label{eq:e9}
\end{equation}
where $f_{ext}$ denotes the feature extraction function and $B_0$ is the extracted feature to be sent to the first memory block.
Supposing $M$ memory blocks are stacked to act as the feature mapping, we have
\begin{equation}\
\begin{aligned}
  B_m = \mathcal{M}_m(B_{m-1}) = \mathcal{M}_m(\mathcal{M}_{m-1}(...(\mathcal{M}_1(B_0))...)),
  \end{aligned}
  \label{eq:e10}
\end{equation}
where $\mathcal{M}_m$ denotes the $m$-th memory block function and $B_{m-1}$ and $B_m$ are the input and output of the $m$-th memory block respectively.
Finally, instead of learning the direct mapping from the low-quality image to the high-quality image, our model uses a convolutional layer in ReconNet to reconstruct the residual image~\cite{VDSR_CVPR16,DRCN_CVPR16,DnCNN}. Therefore, our basic MemNet can be formulated as,
\begin{equation}\
\begin{aligned}
 & \mathbf{y}  =  \mathcal{D}(\mathbf{x}) \\
  & = f_{rec}(\mathcal{M}_M(\mathcal{M}_{M-1}(...(\mathcal{M}_1(f_{ext}(\mathbf{x})))...)))+\mathbf{x},
   \end{aligned}
  \label{eq:e11}
\end{equation}
where $f_{rec}$ denotes the reconstruction function and $ \mathcal{D}$ denotes the function of our basic MemNet.

Given a training set $\{\mathbf{x}^{(i)},\tilde{\mathbf{x}}^{(i)}\}_{i=1}^N$, where $N$ is the number of training patches and $\tilde{\mathbf{x}}^{(i)}$ is the ground truth high-quality patch of the low-quality patch $\mathbf{x}^{(i)}$, the loss function of our basic MemNet with the parameter set $\Theta$, is
\begin{equation}\
  \mathcal{L}(\Theta) = \frac{1}{2N}\sum_{i=1}^{N}\|\tilde{\mathbf{x}}^{(i)} - \mathcal{D}(\mathbf{x}^{(i)})\|^2,
  \label{eq:e12}
\end{equation}\vspace{-2mm}

\begin{figure*}[tbp]
  \centering
  \includegraphics[trim={0 0 0 0mm},clip,width=0.9\linewidth]{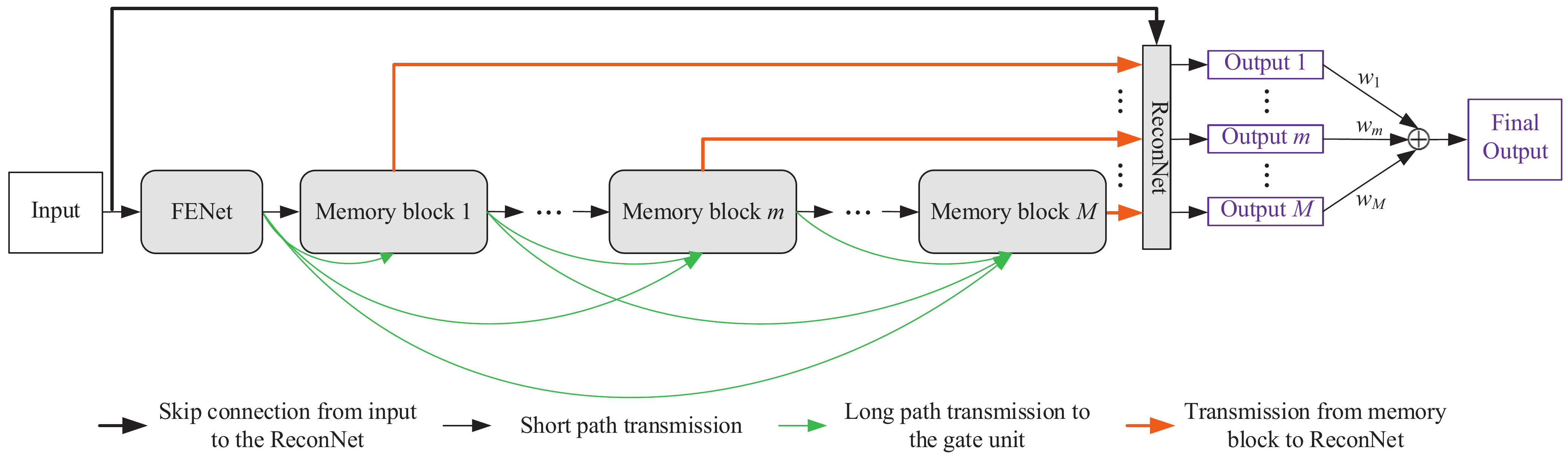}
  \caption{\small Multi-supervised MemNet architecture. The outputs with \textcolor{Purple}{purple} color are supervised. }
  \label{fig:MultiSupervisedNet} \figvspace
\end{figure*}

\SubSection{Memory Block}\label{section:3.2}
We now present the details of our memory block.
The memory block contains a recursive unit and a gate unit.

\Paragraph{Recursive Unit}
is used to model a non-linear function that acts like a recursive synapse in the brain~\cite{Neuroscience,RCNN_CVPR15}.
Here, we use a \textit{residual building block}, which is introduced in ResNet~\cite{ResNet_CVPR16} and shows powerful learning ability for object recognition, as a recursion in the recursive unit.
A residual building block in the $m$-th memory block is formulated as,
\begin{equation}\
  H_m^r = \mathcal{R}_m(H_m^{r-1})=\mathcal{F}(H_m^{r-1},W_m) + H_m^{r-1},
  \label{eq:e1}
\end{equation}
where $H_m^{r-1}$, $H_m^r$ are the input and output of the $r$-th residual building block respectively.
When $r=1$, $H_m^0=B_{m-1}$.
$\mathcal{F}$ denotes the residual function, $W_m$ is the weight set to be learned and $\mathcal{R}$ denotes the function of residual building block.
Specifically, each residual function contains two convolutional layers with the pre-activation structure~\cite{IM_arxiv16},
\begin{equation}\
  \mathcal{F}(H_m^{r-1},W_m)=W_m^2\tau(W_m^1\tau(H_m^{r-1})),
  \label{eq:e2}
\end{equation}
where $\tau$ denotes the activation function, including batch normalization~\cite{BN_ICML15} followed by ReLU~\cite{ReLU_ICML10}, and $W_m^i, i = 1,2$ are the weights of the $i$-th convolutional layer.
The bias terms are omitted for simplicity.

Then, several recursions are recursively learned to generate multi-level representations under different receptive fields.
We call these representations as the short-term memory.
Supposing there are $R$ recursions in the recursive unit, the $r$-th recursion in recursive unit can be formulated as,
\begin{equation}\
  \small
  H_m^r = \mathcal{R}_m^{(r)}(B_{m-1})= \underbrace{\mathcal{R}_m(\mathcal{R}_m(...(\mathcal{R}_m}_r (B_{m-1}))...)),
  \label{eq:e3}
\end{equation}
where $r$-fold operations of $ \mathcal{R}_m$ are performed and $\{H_m^r\}_{r=1}^R$ are the multi-level representations of the recursive unit.
These representations are concatenated as the short-term memory: $B_m^{short} = [ H_m^1, H_m^2, ..., H_m^R]$.
In addition, the long-term memory coming from the previous memory blocks can be constructed as: $B_m^{long} = [ B_{0}, B_{1}, ..., B_{m-1}]$.
The two types of memories are then concatenated as the input to the gate unit,
\begin{equation}\
  B_m^{gate} = [ B_m^{short},  B_m^{long}].
  \label{eq:e6}
\end{equation}

\Paragraph{Gate Unit}
is used to achieve persistent memory through an adaptive learning process. In this paper, we adopt a $1\times1$ convolutional layer to accomplish the \textit{gating mechanism} that can learn adaptive weights for different memories,
\begin{equation}\
  B_m =  f_m^{gate}(B_m^{gate})=W_m^{gate}\tau(B_m^{gate}),
  \label{eq:e7}
\end{equation}
where $f_m^{gate}$ and $B_m$ denote the function of the $1\times1$ convolutional layer (parameterized by $W_m^{gate}$) and the output of the $m$-th memory block, respectively.
As a result, the weights for the long-term memory controls how much of the previous states should be reserved, and the weights for the short-term memory decides how much of the current state should be stored.
Therefore, the formulation of the $m$-th memory block can be written as,
\begin{equation}\
  \begin{aligned}
    &B_m  =  \mathcal{M}_m(B_{m-1}) \\
    &= f_{gate}([\mathcal{R}_m(B_{m-1}), ...,\mathcal{R}_m^{(R)}(B_{m-1}), B_{0}, ..., B_{m-1}]).
  \end{aligned}
  \label{eq:e8}
\end{equation}

\SubSection{Multi-Supervised MemNet}
To further explore the features at different states, inspired by~\cite{DRCN_CVPR16}, we send the output of each memory block to the \textit{same} reconstruction net $\hat{f}_{rec}$  to generate
\begin{equation}\
  \mathbf{y}_m = \hat{f}_{rec}(\mathbf{x},B_m) = \mathbf{x}+f_{rec}(B_m),
  \label{eq:e13}
\end{equation}
where $\{\mathbf{y}_m\}_{m=1}^M$ are the intermediate predictions.
All of the predictions are supervised during training, and used to compute the final output via weighted averaging: $\mathbf{y} = \sum_{m=1}^{M}w_m\cdot \mathbf{y}_m$ (Fig.~\ref{fig:MultiSupervisedNet}).
The optimal weights $\{w_m\}_{m=1}^M$ are automatically learned during training and the final output from the ensemble is also supervised.
The loss function of our multi-supervised MemNet can be formulated as,
\begin{equation}\
  \begin{aligned}
    &\mathcal{L}(\Theta) = \frac{\alpha}{2N}\sum_{i=1}^{N}\|\tilde{\mathbf{x}}^{(i)} - \sum_{m=1}^{M}w_m\cdot \mathbf{y}_m^{(i)}\|^2  \\
    & + \frac{1-\alpha}{2MN}\sum_{m=1}^{M}\sum_{i=1}^{N}\|\tilde{\mathbf{x}}^{(i)} -  \mathbf{y}_m^{(i)}\|^2,
  \end{aligned}
  \label{eq:e15}
\end{equation}
where $\alpha$ denotes the loss weight.

\begin{figure}[t!]
  \centering
  \includegraphics[trim={0 0 0 0mm},clip,width=1\linewidth]{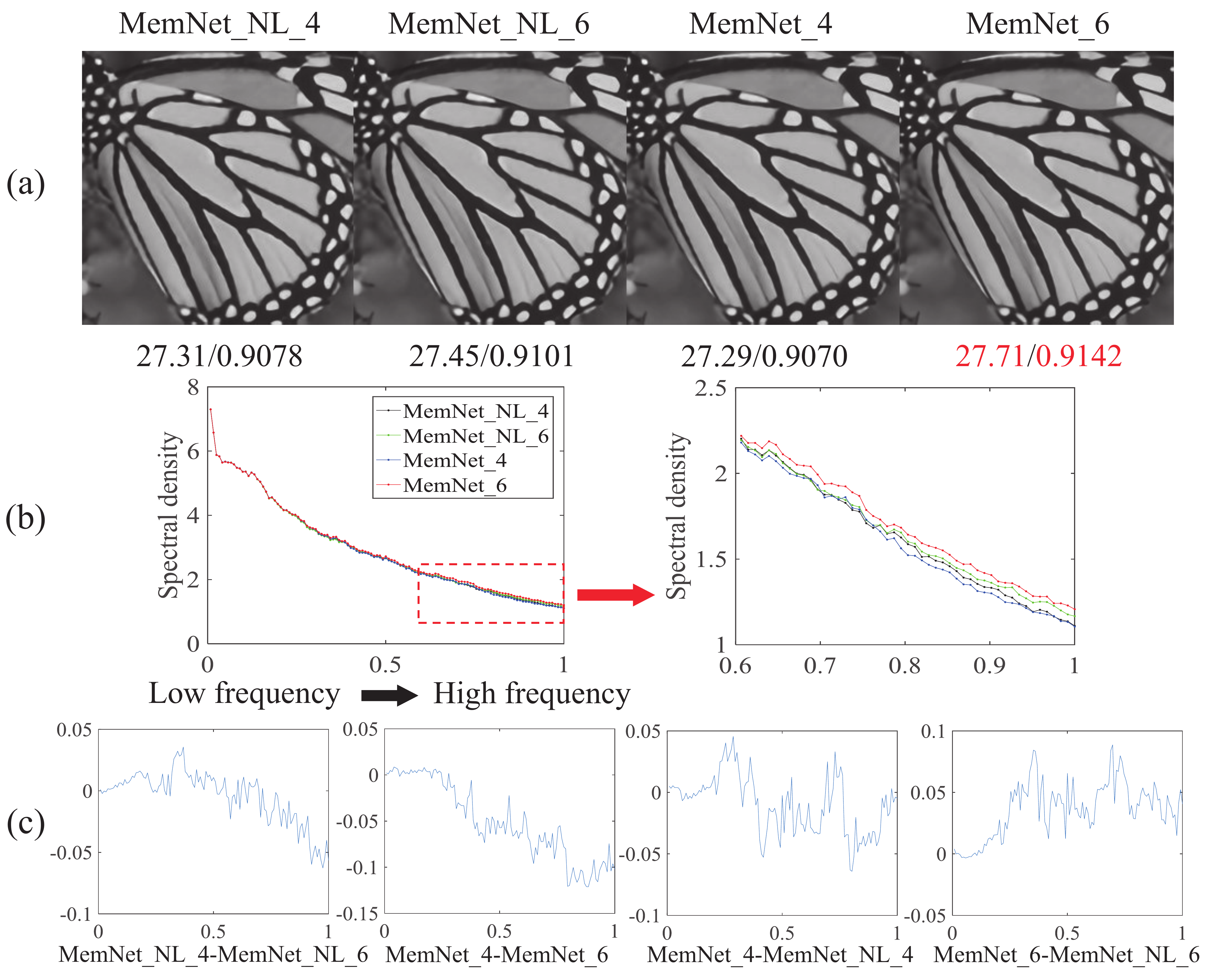}
  \caption{\small (a) $\times4$ super-resolved images and PSNR/SSIMs of different networks.
  (b) We convert $2$-D power spectrums to $1$-D spectral densities by integrating the spectrums along each concentric circle.
  (c) Differences of spectral densities of two networks.}
  \label{fig:R1}\vspace{-6mm}
\end{figure}

\SubSection{Dense Connections for Image Restoration}\label{section:3.4}
Now we analyze why the \textit{long-term dense connections} in MemNet may benefit the image restoration.
In very deep networks, some of the mid/high-frequency information can get lost at latter layers during a typical feedforward CNN process,  and dense connections from previous layers can compensate such loss and further enhance high-frequency signals.
To verify our intuition, we train a $80$-layer MemNet without long-term connections, which is denoted as MemNet$\_$NL, and compare with the original MemNet. 
Both networks have $6$ memory blocks leading to $6$ intermediate outputs, and each memory block contains $6$ recursions.
Fig.~\ref{fig:R1}(a) shows the $4$th and $6$th outputs of both networks.
We compute their power spectrums, center them, estimate spectral densities for a continuous set of frequency ranges from low to high by placing concentric circles, and plot the densities of four outputs in Fig.~\ref{fig:R1}(b).

We further plot {\it differences} of these densities in Fig.~\ref{fig:R1}(c).
From left to right, the first case indicates the earlier layer does contain some mid-frequency information that the latter layers lose.
The $2$nd case verifies that with dense connections, the latter layer absorbs the information carried from the previous layers, and even generate more mid-frequency information. 
The $3$rd case suggests in earlier layers, the frequencies are similar between two models.
The last case demonstrates the MemNet recovers more high frequency  than the version without long-term connections.

\Section{Discussions}\label{section:4}

\Paragraph{Difference to Highway Network}
First, we discuss how the memory block accomplishes the gating mechanism and present the difference between MemNet and Highway Network -- a very deep CNN model using a gate unit to regulate information flow~\cite{Highway}.

To avoid information attenuation in very deep plain networks, inspired by LSTM, Highway Network introduced the bypassing layers along with gate units, i.e., 
\begin{equation}\
  \mathbf{b} = \mathcal{A}(\mathbf{a})\cdot \mathcal{T}(\mathbf{a})+\mathbf{a}\cdot(1-\mathcal{T}(\mathbf{a})),
  \label{eq:e16}
\end{equation}
where $\mathbf{a}$ and $\mathbf{b}$ are the input and output, $\mathcal{A}$ and $\mathcal{T}$ are two non-linear transform functions. $\mathcal{T}$ is the \textit{transform gate} to control how much information produced by $\mathcal{A}$ should be stored to the output; and $1-\mathcal{T}$ is the \textit{carry gate} to decide how much of the input should be reserved to the output.


In MemNet, 
the short-term and long-term memories are concatenated.
The $1\times1$ convolutional layer adaptively learns the weights for different memories.
Compared to Highway Network that learns specific weight for each \textit{pixel}, our gate unit learns specific weight for each \textit{feature map}, which has two advantages: ($1$) to reduce model parameters and complexity; ($2$) to be less prone to overfitting.

\Paragraph{Difference to DRCN}  
There are three main differences between MemNet and DRCN~\cite{DRCN_CVPR16}.
The first is the design of the basic module in network.
In DRCN, the basic module is a \textit{convolutional layer}; while in MemNet, the basic module is a \textit{memory block} to achieve persistent memory.
The second is in DRCN, the weights of the basic modules (i.e., the convolutional layers) are \textit{shared}; while in MemNet, the weights of the memory blocks are \textit{different}.
The third is there are no dense connections among the basic modules in DRCN, which results in a \textit{chain} structure; while in MemNet, there are long-term dense connections among the memory blocks leading to the \textit{multi-path} structure, which not only helps information flow across the network, but also encourages gradient backpropagation during training.
Benefited from the good information flow ability, MemNet could be easily trained without the multi-supervision strategy, which is \textit{imperative} for training DRCN~\cite{DRCN_CVPR16}.

%

\Paragraph{Difference to DenseNet}
Another related work to MemNet is DenseNet~\cite{DenseNet}, which also builds upon a densely connected principle.
In general, DenseNet deals with object recognition, while MemNet is proposed for image restoration.
In addition, DenseNet adopts the densely connected structure in a \textit{local} way (i.e., inside a dense block), while MemNet adopts the densely connected structure in a \textit{global} way (i.e., across the memory blocks).
In Secs.~\ref{section:3.4} and~\ref{section:5.2}, we analyze and demonstrate the long-term dense connections in MemNet indeed play an important role in image restoration tasks.

\begin{table}[t!]
\footnotesize
\centering
\begin{tabular}{|c|c|c|c|c|c|c|c||c|c|}
\hline
Methods  & MemNet$\_$NL & MemNet$\_$NS & MemNet  \\
\hline\hline
$\times2$ & $37.68$/$0.9591$ & $37.71$/$0.9592$ & \textcolor{red}{$37.78$}/\textcolor{red}{$0.9597$} \\
\hline
$\times3$ & $33.96$/$0.9235$ & $34.00$/$0.9239$ & \textcolor{red}{$34.09$}/\textcolor{red}{$0.9248$} \\
\hline
$\times4$ & $31.60$/$0.8878$ & $31.65$/$0.8880$ & \textcolor{red}{$31.74$}/\textcolor{red}{$0.8893$} \\
\hline
\end{tabular}
\caption{\small Ablation study on effects of long-term and short-term connections.
Average PSNR/SSIMs for the scale factor $\times2$, $\times3$ and $\times4$ on dataset Set$5$.
\textcolor{red}{Red} indicates the best performance.}
\label{table:R1} \vspace{-4mm}
\end{table}

\begin{figure}[t!]
  \centering
  \includegraphics[trim={0 0 0 0mm},clip,width=0.95\linewidth]{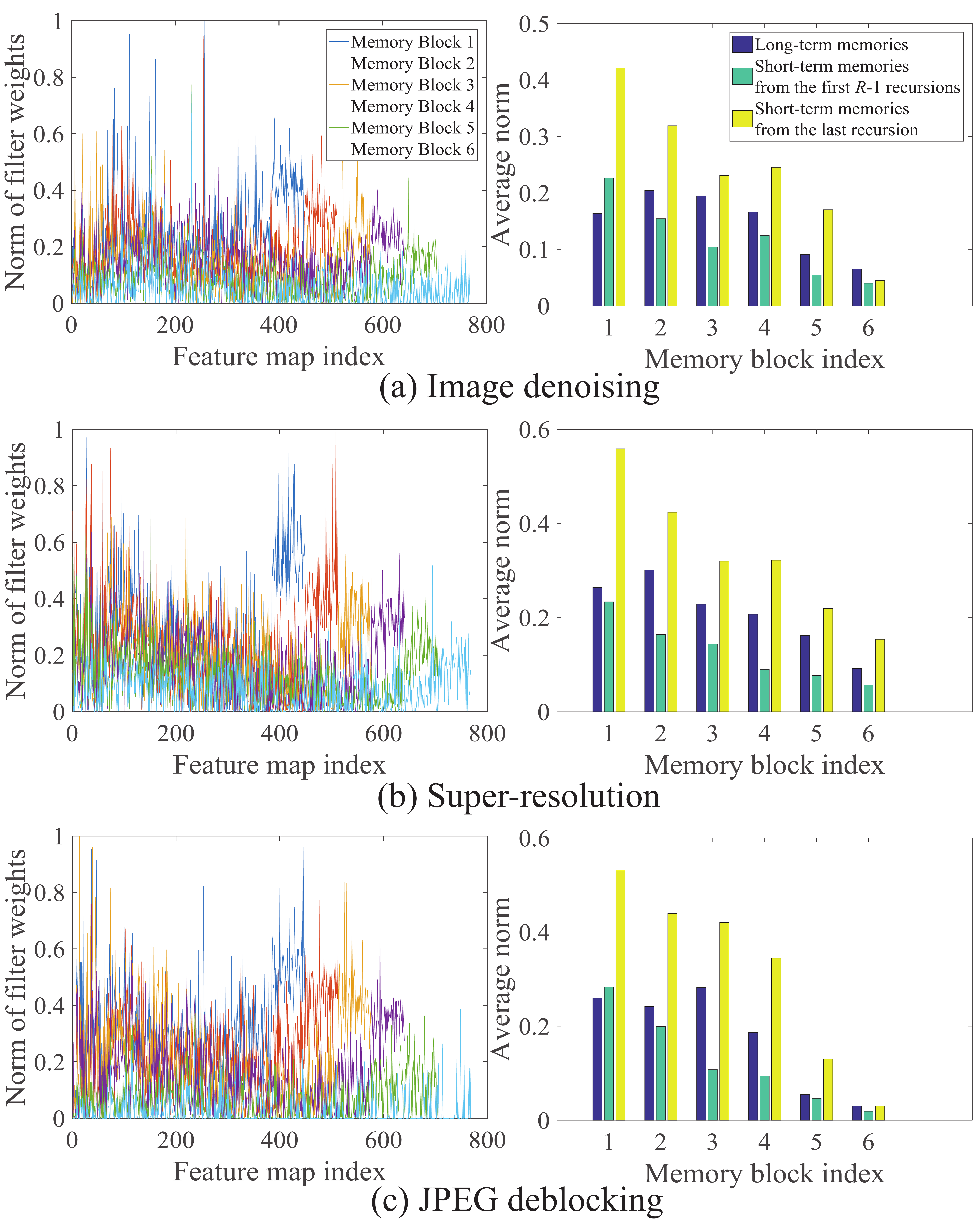}
\caption{\small The norm of filter weights $v_m^l$ vs.~feature map index $l$.
For the curve of the $m$th block, the left ($m\times 64$) elements denote the long-term memories and the rest ($L_m-m\times 64$) elements denote the short-term memories.
The bar diagrams illustrate the average norm of long-term memories, short-term memories from the first $R-1$ recursions and from the last recursion, respectively.
E.g., each yellow bar is the average norm of the short-term memories from the last recursion in the recursive unit (i.e., the last $64$ elements in each curve).
}
\label{fig:gatingparams} \figvspace \vspace{1.5mm}
\end{figure}

\begin{figure*}[t!]
  \begin{minipage}[b]{0.65\textwidth}
	\footnotesize
	\centering
	\begin{tabular}{|c|c|c|c|c|c|c|c|}
		\hline
		Dataset & VDSR~\cite{VDSR_CVPR16} & DRCN~\cite{DRCN_CVPR16} & RED~\cite{RED_NIPS16} & \multicolumn{3}{c|}{MemNet}  \\
		\hline\hline
		Depth  & {$20$} & {$20$} & $30$ & \multicolumn{3}{c|}{$80$}  \\
		\hline
		Filters  & \textcolor{red}{$64$} & $256$ & $128$ & \multicolumn{3}{c|}{\textcolor{red}{$64$}} \\
		\hline
		Parameters  & \textcolor{red}{$665$}K & $1,774$K & $4,131$K & \multicolumn{3}{c|}{$677$K} \\
		\hline
		Traing images  & $291$ & \textcolor{red}{$91$} & $300$ & \textcolor{red}{$91$} & \textcolor{red}{$91$} & $291$ \\
		\hline
		Multi-supervision  & No & Yes & No & No & Yes & Yes \\
		\hline
		PSNR  & $33.66$ & $33.82$ & $33.82$ & $33.92$ & $33.98$ & \textcolor{red}{$34.09$} \\
		\hline
	\end{tabular}
	\tabcaption{\small SISR comparisons with start-of-the-art networks for scale factor $\times3$ on Set5. \textcolor{red}{Red} indicates the fewest number or best performance.}
	\label{table:P_vs_NP_Published} \figvspace\vspace{-0mm}
  \end{minipage}
  \begin{minipage}[b]{0.35\textwidth}
    \centering
    \includegraphics[width=0.85\linewidth]{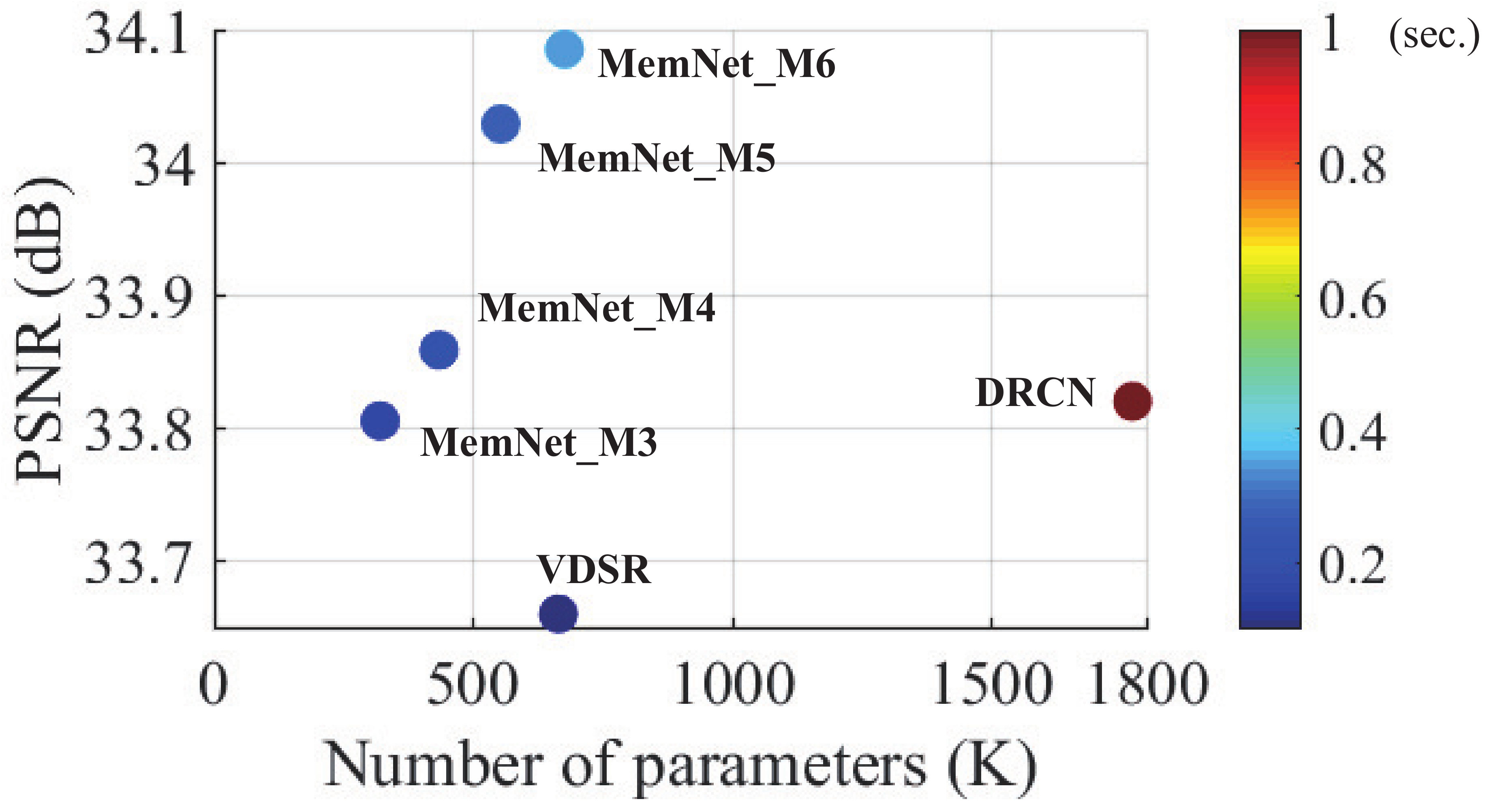}
	\caption{\small PSNR,  complexity vs. speed. 
    }
	\label{fig:psnr_para} \figvspace\vspace{-0mm}
  \end{minipage}%
\end{figure*}

\Section{Experiments}\label{section:5}

\SubSection{Implementation Details}
\Paragraph{Datasets}
For image denoising, we follow~\cite{RED_NIPS16} to use $300$ images from the Berkeley Segmentation Dataset (BSD)~\cite{B200_2001}, known as the train and val sets, to generate image patches as the training set.
Two popular benchmarks, a dataset with $14$ common images and the BSD test set with $200$ images, are used for evaluation.
We generate the input noisy patch by adding Gaussian noise with one of the three noise levels ($\sigma = 30$, $50$ and $70$) to the clean patch.

For SISR, by following the experimental setting in~\cite{VDSR_CVPR16}, we use a training set of $291$ images where $91$ images are from Yang et al.~\cite{ScSR} and other $200$ are from BSD train set.
For testing, four benchmark datasets, Set$5$~\cite{Set5}, Set$14$~\cite{Set14}, BSD$100$\cite{B200_2001} and Urban$100$~\cite{SelfESR_CVPR15} 
are used.
Three scale factors are evaluated, including $\times2$, $\times3$ and $\times4$.
The input LR image is generated by first bicubic downsampling and then bicubic upsampling the HR image with a certain scale.

For JPEG deblocking, the same training set for image denoising is used.
As in~\cite{ARCNN_ICCV15}, Classic$5$ and LIVE$1$ 
are adopted as the test datasets.
Two JPEG quality factors are used, i.e., $10$ and $20$, and the JPEG deblocking input is generated by compressing the image with a certain quality factor using the MATLAB JPEG encoder.

\Paragraph{Training Setting}
Following the method~\cite{RED_NIPS16}, for image denoising, the \textit{grayscale} image is used; while for SISR and JPEG deblocking, the \textit{luminance} component is fed into the model.
The input image size can be arbitrary due to the fully convolution architecture.
Considering both the training time and storage complexities, training images are split into $31 \times 31$ patches with a stride of $21$.
The output of MemNet is the estimated high-quality patch with the same resolution as the input low-quality patch.
We follow~\cite{DRRN} to do data augmentation.
For each task, we train a \textit{single} model for all different levels of corruption.
E.g., for image denoising, noise augmentation is used. Images with different noise levels are all included in the training set.
Similarly, for super-resolution and JPEG deblocking, scale and quality augmentation are used, respectively.

We use Caffe~\cite{Caffe_arxiv14} to implement two $80$-layer MemNet networks, the basic and the multi-supervised versions.
In both architectures, $6$ memory blocks, each contains $6$ recursions, are constructed (i.e., M$6$R$6$).
Specifically, in multi-supervised MemNet, $6$ predictions are generated and used to compute the final output.
$\alpha$ balances different regularizations,
and is empirically set as $\alpha=1/(M+1)$.

The objective functions in Eqn.~\ref{eq:e12} and Eqn.~\ref{eq:e15} are optimized via the mini-batch stochastic gradient descent (SGD) with backpropagation~\cite{CNN_1998}.
We set the mini-batch size of SGD to $64$, momentum parameter to $0.9$, and weight decay to $10^{-4}$.
All convolutional layer has $64$ filters.
Except the $1\times1$ convolutional layers in the gate units, the kernel size of other convolutional layers is $3\times3$.
We use the method in~\cite{MSRA_ICCV15} for weight initialization.
The initial learning rate is set to $0.1$ and then divided $10$ every $20$ epochs.
Training a $80$-layer basic MemNet by $91$ images~\cite{ScSR} for SISR roughly takes $5$ days using $1$ Tesla P$40$ GPU.
Due to space constraint and more recent baselines, we focus on SISR in Sec.~\ref{section:5.2},  ~\ref{section:5.4} and~\ref{section:5.6}, while all three tasks in Sec.~\ref{section:5.3} and~\ref{section:5.5}.

\vspace{-1mm}
\SubSection{Ablation Study}\label{section:5.2}
Tab.~\ref{table:R1} presents the ablation study on the effects of long-term and short-term connections.
Compared to MemNet, MemNet$\_$NL removes the long-term connections (green curves in Fig.~\ref{fig:MultiSupervisedNet}) and MemNet$\_$NS removes the short-term connections (black curves from the first $R-1$ recursions to the gate unit in Fig.~\ref{fig:Differences}.
Connection from the last recursion to the gate unit is \textit{reserved} to avoid a broken interaction between recursive unit and gate unit).
The three networks have the same depth ($80$) and filter number ($64$).
We see that, long-term dense connections are very important since MemNet significantly outperforms MemNet$\_$NL.
Further, MemNet achieves better performance than MemNet$\_$NS, which reveals the short-term connections are also useful for image restoration but less powerful than the long-term connections.
The reason is that the long-term connections skip much more layers than the short-term ones, which can carry some mid/high frequency signals from very early layers to latter layers as described in Sec.~\ref{section:3.4}.

\begin{table*}[t!]
\footnotesize
\centering
  \resizebox{0.85\textwidth}{!}{
\begin{tabular}{|p{1cm}|p{0.5cm}|p{1.5cm}|p{1.5cm}|p{1.5cm}|p{1.5cm}|p{1.5cm}|p{1.5cm}||p{1.6cm}|}
\hline
Dataset & Noise & BM3D~\cite{BM3D} & EPLL~\cite{EPLL}  & PCLR~\cite{PCLR} & PGPD~\cite{PGPD} & WNNM~\cite{WNNM} & RED~\cite{RED_NIPS16} & MemNet \\
\hline\hline
\multirow{3}{*}{$14$ images} & $30$ & $28.49$/$0.8204$ & $28.35$/$0.8200$ & $28.68$/$0.8263$ & $28.55$/$0.8199$ & $28.74$/$0.8273$  & \textcolor{blue}{$29.17$}/\textcolor{blue}{$0.8423$} &  \textcolor{red}{$29.22$}/\textcolor{red}{$0.8444$} \\
                             & $50$ & $26.08$/$0.7427$ & $25.97$/$0.7354$ & $26.29$/$0.7538$ & $26.19$/$0.7442$ & $26.32$/$0.7517$  & \textcolor{blue}{$26.81$}/\textcolor{blue}{$0.7733$} &  \textcolor{red}{$26.91$}/\textcolor{red}{$0.7775$} \\
                             & $70$ & $24.65$/$0.6882$ & $24.47$/$0.6712$ & $24.79$/$0.6997$ & $24.71$/$0.6913$ & $24.80$/$0.6975$  & \textcolor{blue}{$25.31$}/\textcolor{blue}{$0.7206$} &  \textcolor{red}{$25.43$}/\textcolor{red}{$0.7260$} \\
\hline\hline
\multirow{3}{*}{BSD$200$}     & $30$ & $27.31$/$0.7755$ & $27.38$/$0.7825$ & $27.54$/$0.7827$ & $27.33$/$0.7717$ & $27.48$/$0.7807$  & \textcolor{blue}{$27.95$}/\textcolor{blue}{$0.8019$} & \textcolor{red}{$28.04$}/\textcolor{red}{$0.8053$} \\
							  & $50$ & $25.06$/$0.6831$ & $25.17$/$0.6870$ & $25.30$/$0.6947$ & $25.18$/$0.6841$ & $25.26$/$0.6928$  & \textcolor{blue}{$25.75$}/\textcolor{blue}{$0.7167$} & \textcolor{red}{$25.86$}/\textcolor{red}{$0.7202$}\\
							  & $70$ & $23.82$/$0.6240$ & $23.81$/$0.6168$ & $23.94$/$0.6336$ & $23.89$/$0.6245$ & $23.95$/$0.6346$  & \textcolor{blue}{$24.37$}/\textcolor{blue}{$0.6551$} & \textcolor{red}{$24.53$}/\textcolor{red}{$0.6608$} \\
\hline
\end{tabular}
}
\caption{\small Benchmark image denoising results. Average PSNR/SSIMs for noise level $30$, $50$ and $70$ on $14$ images and BSD$200$. \textcolor{red}{Red} color indicates the best performance and \textcolor{blue}{blue} color indicates the second best performance.}
\label{table:Image-denoising}
\end{table*}

\begin{table*}[h!]
\vspace{-2mm}
\footnotesize
\centering
  \resizebox{\textwidth}{!}{
\begin{tabular}{|c|c|c|c|c|c|c|c|c||c|c|}
\hline
Dataset & Scale & Bicubic & SRCNN \cite{SRCNN_PAMI16} & VDSR \cite{VDSR_CVPR16} & DRCN \cite{DRCN_CVPR16} & DnCNN \cite{DnCNN} & LapSRN~\cite{LapSRN} & DRRN~\cite{DRRN}  & MemNet \\
\hline\hline
\multirow{3}{*}{Set$5$}      & $\times2$ & $33.66$/$0.9299$ & $36.66$/$0.9542$ & $37.53$/$0.9587$ & $37.63$/$0.9588$ & $37.58$/$0.9590$ & $37.52$/$0.959$ & \textcolor{blue}{$37.74$}/\textcolor{blue}{$0.9591$} & \textcolor{red}{$37.78$}/\textcolor{red}{$0.9597$} \\
                            & $\times3$ & $30.39$/$0.8682$ & $32.75$/$0.9090$ & $33.66$/$0.9213$ & $33.82$/$0.9226$ & $33.75$/$0.9222$   & $-$/$-$         & \textcolor{blue}{$34.03$}/\textcolor{blue}{$0.9244$} & \textcolor{red}{$34.09$}/\textcolor{red}{$0.9248$} \\
                            & $\times4$ & $28.42$/$0.8104$ & $30.48$/$0.8628$ & $31.35$/$0.8838$ & $31.53$/$0.8854$ & $31.40$/$0.8845$   & $31.54$/$0.885$ & \textcolor{blue}{$31.68$}/\textcolor{blue}{$0.8888$} & \textcolor{red}{$31.74$}/\textcolor{red}{$0.8893$} \\
\hline\hline
\multirow{3}{*}{Set$14$}    & $\times2$ & $30.24$/$0.8688$ & $32.45$/$0.9067$ & $33.03$/$0.9124$ & $33.04$/$0.9118$ & $33.03$/$0.9128$ & $33.08$/$0.913$ & \textcolor{blue}{$33.23$}/\textcolor{blue}{$0.9136$} & \textcolor{red}{$33.28$}/\textcolor{red}{$0.9142$} \\
                            & $\times3$ & $27.55$/$0.7742$ & $29.30$/$0.8215$ & $29.77$/$0.8314$ & $29.76$/$0.8311$ & $29.81$/$0.8321$   & $-$/$-$         & \textcolor{blue}{$29.96$}/\textcolor{blue}{$0.8349$} & \textcolor{red}{$30.00$}/\textcolor{red}{$0.8350$}  \\
                            & $\times4$ & $26.00$/$0.7027$ & $27.50$/$0.7513$ & $28.01$/$0.7674$ & $28.02$/$0.7670$ & $28.04$/$0.7672$   & $28.19$/$0.772$ & \textcolor{blue}{$28.21$}/\textcolor{blue}{$0.7721$} & \textcolor{red}{$28.26$}/\textcolor{red}{$0.7723$} \\
\hline\hline
\multirow{3}{*}{BSD$100$}   & $\times2$ & $29.56$/$0.8431$ & $31.36$/$0.8879$ & $31.90$/$0.8960$ & $31.85$/$0.8942$ & $31.90$/$0.8961$ & $31.80$/$0.895$ & \textcolor{blue}{$32.05$}/\textcolor{blue}{$0.8973$} & \textcolor{red}{$32.08$}/\textcolor{red}{$0.8978$} \\
                            & $\times3$ & $27.21$/$0.7385$ & $28.41$/$0.7863$ & $28.82$/$0.7976$ & $28.80$/$0.7963$ & $28.85$/$0.7981$                       & $-$/$-$         & \textcolor{blue}{$28.95$}/\textcolor{red}{$0.8004$} & \textcolor{red}{$28.96$}/\textcolor{blue}{$0.8001$} \\
                            & $\times4$ & $25.96$/$0.6675$ & $26.90$/$0.7101$ & $27.29$/$0.7251$ & $27.23$/$0.7233$ & $27.29$/$0.7253$     & $27.32$/$0.728$ & \textcolor{blue}{$27.38$}/\textcolor{red}{$0.7284$} & \textcolor{red}{$27.40$}/\textcolor{blue}{$0.7281$} \\
\hline\hline
\multirow{3}{*}{Urban$100$} & $\times2$ & $26.88$/$0.8403$ & $29.50$/$0.8946$ & $30.76$/$0.9140$ & $30.75$/$0.9133$ & $30.74$/$0.9139$ 				   & $30.41$/$0.910$ & \textcolor{blue}{$31.23$}/\textcolor{blue}{$0.9188$} & \textcolor{red}{$31.31$}/\textcolor{red}{$0.9195$} \\
                            & $\times3$ & $24.46$/$0.7349$ & $26.24$/$0.7989$ & $27.14$/$0.8279$ & $27.15$/$0.8276$ & $27.15$/$0.8276$  & $-$/$-$         & \textcolor{blue}{$27.53$}/\textcolor{red}{$0.8378$} & \textcolor{red}{$27.56$}/\textcolor{blue}{$0.8376$} \\
                            & $\times4$ & $23.14$/$0.6577$ & $24.52$/$0.7221$ & $25.18$/$0.7524$ & $25.14$/$0.7510$ & $25.20$/$0.7521$                    & $25.21$/$0.756$ & \textcolor{blue}{$25.44$}/\textcolor{red}{$0.7638$} & \textcolor{red}{$25.50$}/\textcolor{blue}{$0.7630$} \\
\hline
\end{tabular}
}
\caption{\small Benchmark SISR results. Average PSNR/SSIMs for scale factor $\times2$, $\times3$ and $\times4$ on datasets Set$5$, Set$14$, BSD$100$ and Urban$100$.
}
\label{table:Super-resolution}
\end{table*}

\begin{table*}[h!]
\vspace{-2mm}
\footnotesize
\centering
  \resizebox{0.7\textwidth}{!}{
\begin{tabular}{|c|c|c|c|c|c|c|c||c|c|}
\hline
Dataset & Quality & JPEG & ARCNN~\cite{ARCNN_ICCV15} & TNRD~\cite{TNRD} & DnCNN~\cite{DnCNN} & MemNet \\
\hline\hline
\multirow{2}{*}{Classic5} & $10$ & $27.82$/$0.7595$ & $29.03$/$0.7929$ & $29.28$/$0.7992$ & \textcolor{blue}{$29.40$}/\textcolor{blue}{$0.8026$} & \textcolor{red}{$29.69$}/\textcolor{red}{$0.8107$} \\
                          & $20$ & $30.12$/$0.8344$ & $31.15$/$0.8517$ & $31.47$/$0.8576$ & \textcolor{blue}{$31.63$}/\textcolor{blue}{$0.8610$} & \textcolor{red}{$31.90$}/\textcolor{red}{$0.8658$} \\
\hline\hline
\multirow{2}{*}{LIVE1}    & $10$ & $27.77$/$0.7730$ & $28.96$/$0.8076$ & $29.15$/$0.8111$ & \textcolor{blue}{$29.19$}/\textcolor{blue}{$0.8123$} & \textcolor{red}{$29.45$}/\textcolor{red}{$0.8193$} \\
                          & $20$ & $30.07$/$0.8512$ & $31.29$/$0.8733$ & $31.46$/$0.8769$ & \textcolor{blue}{$31.59$}/\textcolor{blue}{$0.8802$} & \textcolor{red}{$31.83$}/\textcolor{red}{$0.8846$} \\
\hline
\end{tabular}
}
\caption{\small Benchmark JPEG deblocking results. Average PSNR/SSIMs for quality factor $10$ and $20$ on datasets Classic$5$ and LIVE$1$.
}
\label{table:JPEG-deblocking}
\vspace{-4mm}
\end{table*}

\SubSection{Gate Unit Analysis}\label{section:5.3}
We now illustrate how our gate unit affects different kinds of memories.
Inspired by~\cite{DenseNet}, we adopt a weight norm as an approximate for the dependency of the current layer on its preceding layers, which is calculated by the corresponding weights from all filters w.r.t.~each feature map: $v_m^l = \sqrt{\sum_{i=1}^{64} (W_m^{gate}(1,1,l,i))^2},\ l=1,2,...,L_m$,
where $L_m$ is the number of the input feature maps for the $m$-th gate unit, $l$ denotes the feature map index, $W_m^{gate}$ stores the weights with the size of $1\times1\times{L_m}\times64$, and $v_m^l$ is the weight norm of the $l$-th feature map for the $m$-th gate unit.
Basically, the larger the norm is, the stronger dependency it has on this particular feature map.
For better visualization, we normalize the norms to the range of $0$ to $1$.
Fig.~\ref{fig:gatingparams} presents the norm of the filter weights $\{v_m^l\}_{m=1}^6$ vs.~feature map index $l$.
We have three observations:
($1$) Different tasks have different norm distributions.
($2$) The average and variance of the weight norms become smaller as the memory block number increases.
($3$) In general, the short-term memories from the last recursion in recursive unit (the last $64$ elements in each curve) contribute most than the other two memories, and the long-term memories seem to play a more important role in late memory blocks to recover useful signals than the short-term memories from the first $R-1$ recursions.

\SubSection{Comparision with Non-Persistent CNN Models}\label{section:5.4}
In this subsection, we compare MemNet with three existing non-persistent CNN models, i.e., VDSR~\cite{VDSR_CVPR16}, DRCN~\cite{DRCN_CVPR16} and RED~\cite{RED_NIPS16}, to demonstrate the superiority of our persistent memory structure.
VDSR and DRCN are two representative networks with the plain structure and RED is representative for skip connections.
Tab.~\ref{table:P_vs_NP_Published} presents the published results of these models along with their training details.
Since the training details are different among different work, we choose DRCN as a baseline, which achieves good performance using the least training images.
But, unlike DRCN that widens its network to increase the parameters (filter number: $256$ vs.~$64$), we deepen our MemNet by stacking more memory blocks (depth: $20$ vs.~$80$).
It can be seen that, using the fewest training images ($91$), filter number ($64$) and relatively few model parameters ($667$K), our basic MemNet
already achieves higher PSNR than the prior networks.
Keeping the setting unchanged, our multi-supervised MemNet further improves the performance.
With more training images ($291$), our MemNet significantly outperforms the state of the arts.

Since we aim to address the long-term dependency problem in networks, we intend to make our MemNet \textit{very deep}.
However, MemNet is also able to balance the model complexity and accuracy.
Fig.~\ref{fig:psnr_para} presents the PSNR of different intermediate predictions in MemNet (e.g., MemNet$\_$M$3$ denotes the prediction of the $3$rd memory block) for scale $\times 3$ on Set$5$, in which the colorbar indicates the inference time (sec.) when processing a $288 \times 288$ image on GPU P$40$.
Results of VDSR~\cite{VDSR_CVPR16} and DRCN~\cite{DRCN_CVPR16} are cited from their papers.
RED~\cite{RED_NIPS16} is skipped here since its high number of parameters may reduce the contrast among other methods.
We see that our MemNet already achieve comparable result at the $3$rd prediction using much fewer parameters, and significantly outperforms the state of the arts by slightly increasing model complexity.

\SubSection{Comparisons with State-of-the-Art Models}\label{section:5.5}
We compare multi-supervised $80$-layer MemNet with the state of the arts in three restoration tasks, respectively.

\begin{figure}[t!]
  \centering
  \includegraphics[trim={0 0 0 0mm},clip,width=1\linewidth]{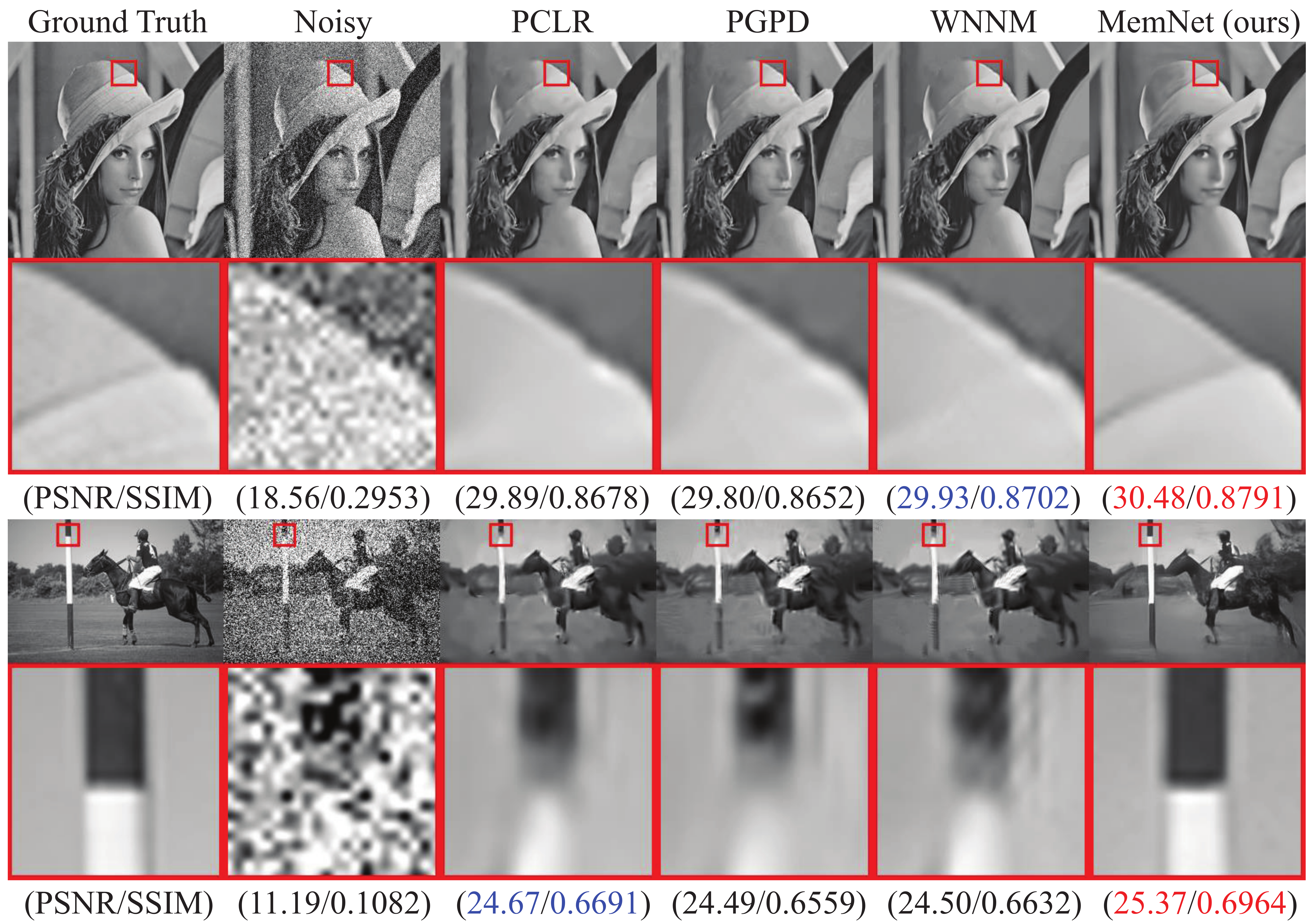}
  \caption{\small Qualitative comparisons of image denoising. The first row shows image ``$10$" from $14$-image dataset with noise level $30$. Only MemNet recovers the fold. The second row shows image ``$206062$" from BSD$200$ with noise level $70$. Only MemNet correctly recovers the pillar. Please zoom in to see the details. }
  \label{fig:final_GD} \figvspace\vspace{0mm}
\end{figure}

\Paragraph{Image Denoising}
Tab.~\ref{table:Image-denoising} presents quantitative results on two benchmarks, with results cited from~\cite{RED_NIPS16}.
For BSD$200$ dataset, by following the setting in RED\cite{RED_NIPS16}, the original image is resized to its half size.
As we can see, our MemNet achieves the best performance on all cases.
It should be noted that, for each test image, RED rotates and mirror flips the kernels, and performs inference multiple times.
The outputs are then averaged to obtain the final result. 
They claimed this strategy can lead to better performance.
However, in our MemNet, we \textit{do not} perform any post-processing.
For qualitative comparisons, we use public codes of PCLR~\cite{PCLR}, PGPD~\cite{PGPD} and WNNM~\cite{WNNM}.
The results are shown in Fig.~\ref{fig:final_GD}.
As we can see, our MemNet handles Gaussian noise better than the previous state of the arts.

\begin{figure}[t!]
  \centering
  \includegraphics[trim={0 0 0 0mm},clip,width=1\linewidth]{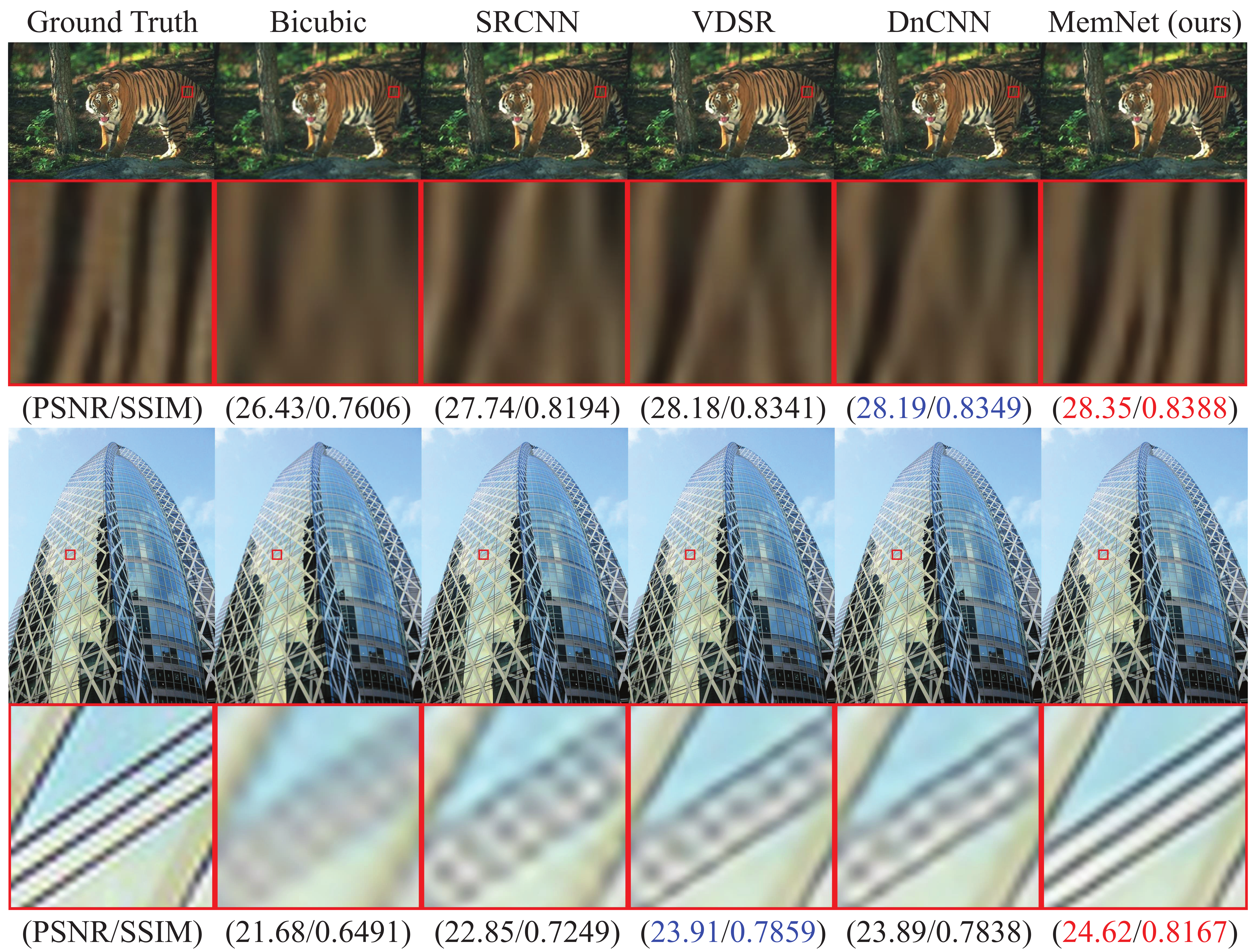}
  \caption{\small Qualitative comparisons of SISR. The first row shows image ``$108005$" from BSD$100$ with scale factor $\times3$. Only MemNet correctly recovers the pattern. The second row shows image ``img$\_002$" from Urban$100$ with scale factor $\times4$. MemNet recovers sharper lines.}
  \label{fig:final_SR} \figvspace \vspace{0.5mm}
\end{figure}

\begin{figure}[t!]
  \centering
  \includegraphics[trim={0 0 0 0mm},clip,width=1\linewidth]{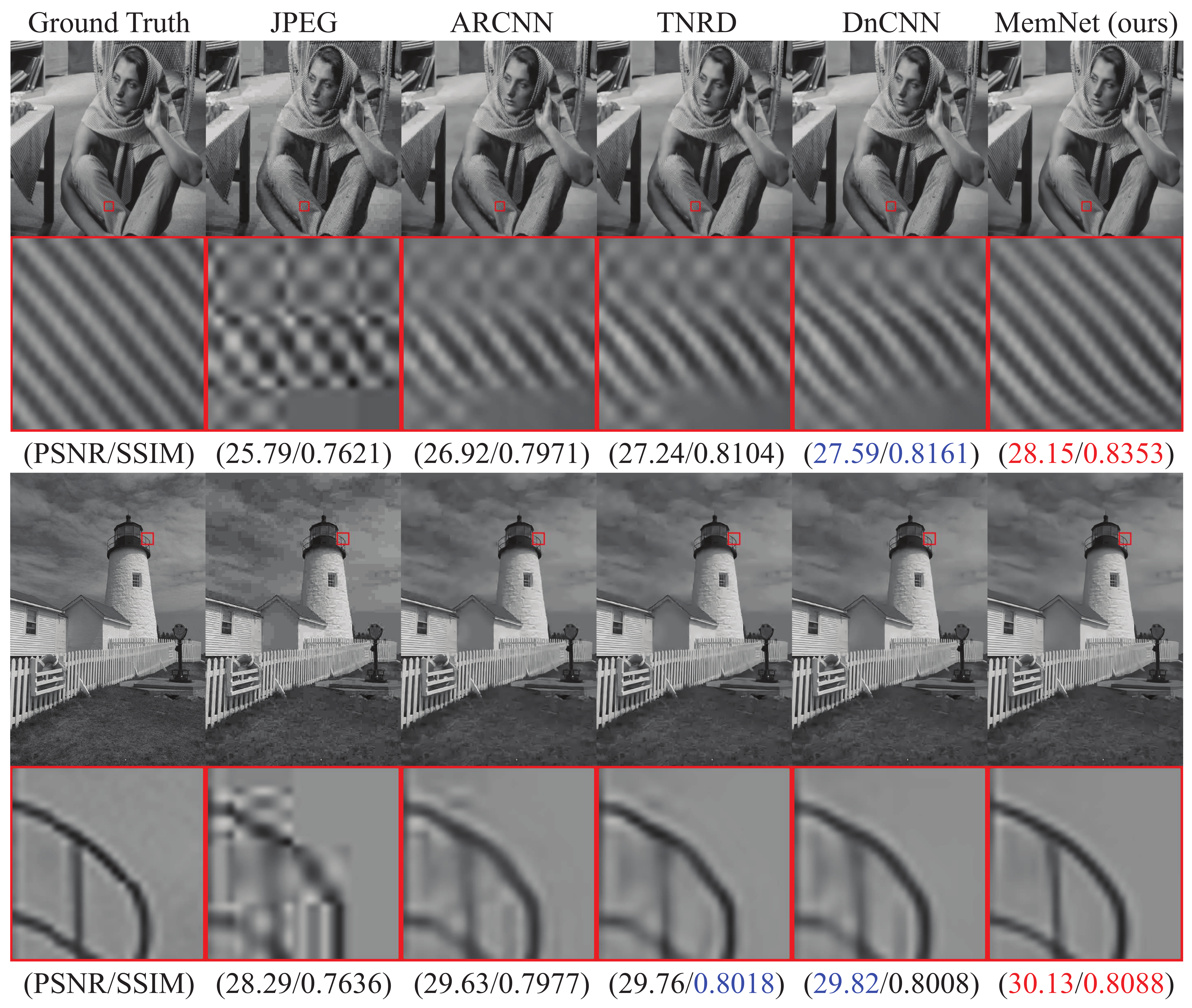}
  \caption{\small Qualitative comparisons of JPEG deblocking. The first row shows image ``barbara" from Classic$5$ with quality factor $10$. MemNet recovers the lines, while others give blurry results. The second row shows image ``lighthouse" from LIVE$1$ with quality factor $10$. MemNet accurately removes the blocking artifact.}
  \label{fig:final_JD} \figvspace \vspace{0mm}
\end{figure}

\Paragraph{Super-Resolution}
Tab.~\ref{table:Super-resolution} summarizes quantitative results on four benchmarks, by citing the results of prior methods. 
MemNet outperforms prior methods in almost all cases. 
Since LapSRN doesn't report the results on scale $\times 3$, we use the symbol '$-$' instead.
Fig.~\ref{fig:final_SR} shows the visual comparisons for SISR.
SRCNN~\cite{SRCNN_PAMI16}, VDSR~\cite{VDSR_CVPR16} and DnCNN~\cite{DnCNN} are compared using their public codes.
MemNet recovers relatively sharper edges, while others have blurry results.

\Paragraph{JPEG Deblocking}
Tab.~\ref{table:JPEG-deblocking} shows the JPEG deblocking results on Classic$5$ and LIVE$1$, by citing the results from~\cite{DnCNN}.
Our network significantly outperforms the other methods, and deeper networks do improve the performance compared to the shallow one, e.g., ARCNN.
Fig.~\ref{fig:final_JD} shows the JPEG deblocking results of these three methods, which are generated by their corresponding public codes.
As it can be seen, MemNet effectively removes the blocking artifact and recovers higher quality images than the previous methods.


\SubSection{Comparison on Different Network Depths}\label{section:5.6}
Finally, we present the comparison on different network depths, which is caused by stacking different numbers of memory blocks or recursions.
Specifically, we test four network structures: M$4$R$6$, M$6$R$6$, M$6$R$8$ and M$10$R$10$, which have the depth $54$, $80$, $104$ and $212$, respectively.
Tab.~\ref{table:depths} shows the SISR performance of these networks on Set$5$ with scale factor $\times 3$.
It verifies \textit{deeper is still better} and the proposed deepest network M$10$R$10$ achieves \textbf{$34.23$} dB, with the improvement of $0.14$ dB compared to M$6$R$6$.

\begin{table}[t!]
\vspace{3mm}
\small
\centering
\begin{tabular}{|c|c|c|c|c|c|c|c||c|c|}
\hline
Network  & M$4$R$6$ & M$6$R$6$ & M$6$R$8$ & M$10$R$10$ \\
\hline
Depth  & $54$ & $80$ & $104$ & $212$ \\
\hline
PSNR (dB) & $34.05$ & $34.09$ & $34.16$ & \textcolor{red}{$34.23$} \\
\hline
\end{tabular}
\caption{\small Comparison on different network depths.}
\label{table:depths} \figvspace\vspace{-1mm}
\end{table}

\vspace{-2mm}
\section{Conclusions}

In this paper, a very deep end-to-end persistent memory network (MemNet) is proposed for image restoration, where a memory block accomplishes the gating mechanism for tackling the long-term dependency problem in the previous CNN architectures.
In each memory block, a recursive unit is adopted to learn multi-level representations as the short-term memory.
Both the short-term memory from the recursive unit and the long-term memories from the previous memory blocks are sent to a gate unit, which adaptively learns different weights for different memories.
We use the same MemNet structure to handle image denoising, super-resolution and JPEG deblocking simultaneously.
Comprehensive benchmark evaluations well demonstrate the superiority of our MemNet over the state of the arts.

{\small
\bibliographystyle{ieee}
\bibliography{egbib}
}

\end{document}